\documentclass[runningheads]{llncs}
\usepackage{graphicx}
\usepackage{hyperref}
\hypersetup{hidelinks}
\usepackage{tikz}
\usepackage{comment}
\usepackage{amsmath,amssymb} 
\usepackage{color}
\usepackage{multirow}
\usepackage{marvosym}

\usepackage{orcidlink}

\usepackage[accsupp]{axessibility}  

\begin{document}
\pagestyle{headings}
\mainmatter

\title{PASS: Part-Aware Self-Supervised Pre-Training for Person Re-Identification} 

\titlerunning{PASS: Part-Aware Self-Supervised Pre-Training for Person Re-Identification}
%
\author{Kuan Zhu\inst{1,2}\orcidlink{0000-0002-1670-944X} \and
Haiyun Guo\inst{1,2}\textsuperscript{(\Letter)}\orcidlink{0000-0001-9241-6211} \and
Tianyi Yan\inst{1,2}\orcidlink{0000-0003-3685-2566} \and
Yousong Zhu\inst{1}\orcidlink{0000-0001-8544-410X} \and\\
Jinqiao Wang\inst{1,2,3}\orcidlink{0000-0002-9118-2780} \and
Ming Tang\inst{1,2}\orcidlink{0000-0003-4976-3095}}
\authorrunning{Zhu et al.}
%
\institute{National Laboratory of Pattern Recognition, Institute of Automation,\\Chinese Academy of Sciences \and
School of Artificial Intelligence, University of Chinese Academy of Sciences
 \and
 Peng Cheng Laboratory \\
\email{\{kuan.zhu, haiyun.guo, tianyi.yan, yousong.zhu, jqwang, tangm\} @nlpr.ia.ac.cn}}
\maketitle

\begin{abstract}
In person re-identification (ReID), very recent researches have validated pre-training the models on unlabelled person images is much better than on ImageNet. However, these researches directly apply the existing self-supervised learning (SSL) methods designed for image classification to ReID without any adaption in the framework. These SSL methods match the outputs of $local$ views (e.g., red T-shirt, blue shorts) to those of the $global$ views at the same time, losing lots of details. In this paper, we propose a ReID-specific pre-training method, Part-Aware Self-Supervised pre-training (PASS), which can generate part-level features to offer fine-grained information and is more suitable for ReID. PASS divides the images into several local areas, and the $local$ views randomly cropped from each area are assigned a specific learnable \texttt{[PART]} token. On the other hand, the \texttt{[PART]}s of all local areas are also appended to the $global$ views. PASS learns to match the outputs of the $local$ views and $global$ views on the same \texttt{[PART]}. That is, the learned \texttt{[PART]} of the $local$ views from a local area is only matched with the corresponding \texttt{[PART]} learned from the $global$ views. As a result, each \texttt{[PART]} can focus on a specific local area of the image and extracts fine-grained information of this area. Experiments show PASS sets the new state-of-the-art performances on Market1501 and MSMT17 on various ReID tasks, e.g., vanilla ViT-S/16 pre-trained by PASS achieves 92.2\%/90.2\%/88.5\% mAP accuracy on Market1501 for supervised/UDA/USL ReID. Our codes are available at \textcolor{blue}{\url{https://github.com/CASIA-IVA-Lab/PASS-reID}}.



\end{abstract}

\section{Introduction}

Person re-identification (ReID) aims to associate the person images captured by different cameras. Limited by the scale of the labeled ReID datasets, most existing methods first pre-train the backbone networks (e.g., ResNet \cite{ResNet}, ViT \cite{ViT}) on ImageNet \cite{imagenet} and then fine-tune them on person ReID datasets to boost the performance \cite{PCB,BOT,fastreid}. However, it is arguable whether using ImageNet for pre-training is optimal as there exist large domain gaps between ImageNet and person ReID data, e.g., 1) ImageNet-1K contains a thousand kinds of objects  while ReID datasets only contain persons; 2) the model pre-trained on ImageNet will focus on category-level differences, losing lots of rich visual information. Therefore, the fine-grained identity information, which is preferred by ReID, can not be provided by pre-training on ImageNet \cite{LUPerson,reid_ssl}. 


\begin{figure}[t]
\centering
\includegraphics[width=1.\linewidth]{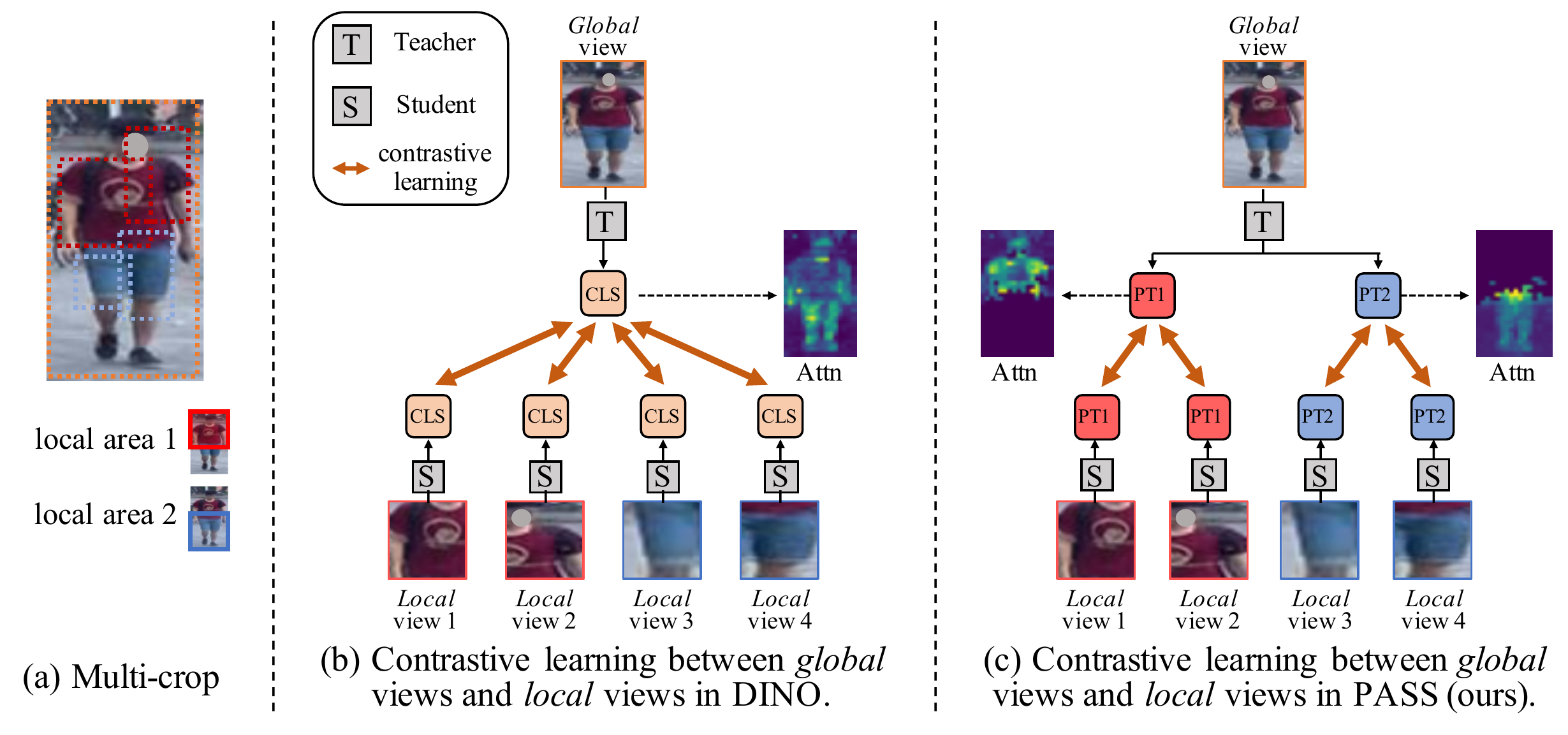}
\caption{(a) Multi-crop \cite{SwAV} is a data augmentation strategy which uses a mix of views with different resolutions. (b) DINO \cite{DINO} learns to match the outputs of all the $local$ views and $global$ views, losing lots of details. (c) In PASS, the $local$ views cropped from each local area are assigned a specific \texttt{[PART]}, e.g., the $local$ views cropped from the upper local area, $local$ view 1 and 2, are assigned \texttt{[PART]}$_1$; the $local$ views cropped from the lower local area, $local$ view 3 and 4, are assigned \texttt{[PART]}$_2$. The $global$ views are appended with all the \texttt{[PART]}s. PASS learns to match the predictions on the same  \texttt{[PART]} (i.e., \texttt{[PART]}$_1$ or \texttt{[PART]}$_2$) of $local$ views and $global$ views, \textbf{respectively}  
}
\label{fig:introduction}
\end{figure}

To bridge the gap between pre-training and fine-tuning datasets for better ReID models, Fu et al. \cite{LUPerson} propose the first large scale unlabeled person ReID dataset ``LUPerson" and demonstrate that unsupervised pre-training the models on LUPerson is quite better than supervised pre-training on ImageNet. Luo et al. \cite{reid_ssl} further investigate that DINO \cite{DINO} algorithm with Transformer architecture (ViT \cite{ViT}) obtains the best ReID performance among the existing self-supervised learning (SSL) methods and network architectures. However, these works \cite{LUPerson,reid_ssl} directly apply the existing SSL methods, that are proposed for image classification, to ReID and do not make any adaption in the SSL framework. Another gap will appear when employing these SSL methods to ReID, which is shown in Figure~\ref{fig:introduction}(b). DINO matches the outputs of all the $local$ views with the $global$ views in the same feature space. It is unreasonable to make the outputs of red T-shirt and blue shorts both match the output of the global image at the same time. To compromise on this matching, only the shared features of different views are retrained and lots of local detailed information are removed in the learned features. However, as fine-grained information has been verified to be crucial to describe a person image \cite{PCB,MGN,SPReID}, it is necessary to design a pre-training method that can extract as many fine-grained clues as possible for ReID. 

In this paper, we propose a ReID-specific pre-training method based on Transformer, Part-Aware Self-Supervised pre-training (PASS), by which part-level features can be automatically extracted to offer fine-grained information for person ReID. PASS uses the learning paradigm of knowledge distillation to match the outputs of the teacher network and student network. It first divides the image into several fixed overlapping local areas and randomly crops $local$ views from these local areas. The $global$ views are randomly cropped from the whole image with higher resolution. All views are passed through the student while only $global$ views are passed through the teacher. For simplicity, we only illustrate the comparison between the outputs of $local$ views passed through student and  $global$ views passed through teacher in Figure~\ref{fig:introduction}(c). Before passing through the student, the $local$ views cropped from each local area are assigned a specific learnable \texttt{[PART]} token, which is used to learn the local representation. All these \texttt{[PART]}s are also appended to the $global$ views and fed to the teacher to learn local features from the whole image. 
PASS learns to match the corresponding \texttt{[PART]}s of $local$ views and $global$ views, where the \texttt{[PART]}s of different areas are not compared.
Take Figure~\ref{fig:introduction}(c) as an example, the $local$ views cropped from the upper local area, $local$ views 1 and  2, are assigned \texttt{[PART]}$_1$; the $local$ views cropped from the lower local area, $local$ views 3 and  4, are assigned \texttt{[PART]}$_2$. The predictions on the \texttt{[PART]}$_1$s of $local$ views 1 and 2 are only compared with those on the \texttt{[PART]}$_1$s of the $global$ views and so does the \texttt{[PART]}$_2$s.


In the student, the $local$ views assigned to each \texttt{[PART]} are cropped from a specific local area, thus the \texttt{[PART]}s can focus on different areas. PASS uses the student to update the teacher, which can guarantee \texttt{[PART]}s in teacher also focus on different local areas and learns fine-grained information.
In pre-training, the student learns to match the output of the teacher on the same \texttt{[PART]},
which can guarantee each \texttt{[PART]} learns a robust local representation from the $local$ views cropped in its corresponding local area. 
In fine-tuning, all \texttt{[PART]}s are appended to the input image and each \texttt{[PART]} automatically learns the local representation for a specific area.


We summarize the contributions of this work as: (i) In this paper, we propose the ReID-specific pre-training method, Part-Aware Self-Supervised pre-training (PASS), which is more suitable for ReID with part-level features offering fine-grained information. It is worth noting that we do not add any complex module to extract part-level features but only use several learnable tokens. (ii) The pre-trained ViT backbone can be fine-tuned on various ReID downstream tasks, i.e., supervised learning, unsupervised domain adaptation (UDA), and unsupervised learning (USL). PASS helps the ViTs set the new state-of-the-art performance on Market-1501 \cite{Market1501} and MSMT17 \cite{MSMT17}, e.g., vanilla ViT-S pre-trained by PASS achieves 92.2\%/90.2\%/88.5\% Rank-1 and 69.1\%/49.1\%/41.0\% mAP accuracy on Market1501 and MSMT17 for supervised/UDA/USL ReID, respectively.


\section{Related Work}

\subsection{Self-Supervised Learning}

Self-supervised learning (SSL) methods aim to learn discriminative representations from large scale unlabeled data. Recently, contrastive learning methods have made remarkable achievements \cite{MoCo_v1,MoCo_v2,MoCo_v3,BYOL,DINO,SwAV,MoBY}, significantly reducing the gap with supervised pre-training. MoCo series \cite{MoCo_v2,MoCo_v3}, which are developed from Momentum Contrast, treat the augmentations of a sample as positive pairs and all other samples as negative pairs. Ge et al. \cite{BYOL} propose a new paradigm, BYOL, where the online network predicts the output representation of the target network on the same image under a different augmented view, removing the need of negative pairs. DINO \cite{DINO} further improves BYOL by using a centering and sharpening of the momentum teacher outputs to avoid model collapse. Besides, DINO adopts the augmentation strategy of multi-crop  \cite{SwAV} to conduct the comparison between the representations of $global$ views and $local$ views. Xie et al. \cite{MoBY} propose MoBY which is a Transformer-specific method and combines MoCo with BYOL. Among these methods, DINO algorithm, plus the ViT architecture, can achieve the best performance on ReID  \cite{reid_ssl}. Therefore, we aim to improve DINO to better adapt to the ReID tasks and obtain higher performance.

\subsection{Person Re-Identification}

\subsubsection{Part-based person ReID.}
Employing part-level features for person image description can offer fine-grained information and has been verified as beneficial for person re-identification. Many efforts have been made to develop the part-based person re-ID to boost the state-of-the-art performance \cite{PCB,MGN,ISP,AAformer,SCS+,CascadeAtt}. PCB \cite{PCB} first directly partitions the person images into fixed horizontal stripes and extract stripe-based part features. MGN \cite{MGN} enhances the robustness by dividing images into stripes of different granularities and designs overlap between stripes. SPReID \cite{SPReID} uses a pre-trained human semantic parsing model to provide the mask of body parts to extract part features. ISP \cite{ISP} proposes to automatically locate both human parts and non-human ones at pixel-level by iterative clustering. TransReID \cite{transreid}, which is a Transformer-specific method, also designs to extract the part-level features by re-arranging the patch embeddings and re-grouping them. All these methods show extracting part-level features to offer fine-grained information is of vital importance for person ReID.

\subsubsection{Unsupervised pre-training for person ReID.}
Recently, more and more works find that there exists large domain gaps between ImageNet and person ReID data \cite{LUPerson,reid_ssl}, which makes pre-training ReID models on unlabelled person images is much better than pre-training on ImageNet. Fu et al. \cite{LUPerson} first focus on this problem and propose a large scale unlabeled person ReID dataset ``LUPerson''. They also validate that unsupervised pre-training on LUPerson can improve the ReID performance compared with ImageNet-1k pre-training. Luo et al. \cite{reid_ssl} further investigate several self-supervised learning methods and backbone networks. The results show that DINO \cite{DINO} with ViT \cite{ViT} obtains the best performance. \cite{reid_ssl} also proposes a data filtering method and IBN-based convolution stem for ViT architecture. However, apart from some adaptions in data augmentation or hyper-parameters, they do not specifically improve these pre-training methods for person ReID. The gaps will appear when applying the existing SSL methods proposed for image classification to ReID as shown in Fig.~\ref{fig:introduction}. In this paper, we propose a ReID-specific self-supervised pre-training method, PASS, which makes the model learn to extract part-level features in pre-training and automatically extract part-level features in fine-tuning.

\section{Method}

\subsection{Preliminaries}

\subsubsection{Vision Transformer.}
We briefly describe the mechanism of the Vision Transformer (ViT) \cite{ViT,Transformer} here, and please refer to \cite{Transformer} for details about Transformers and to \cite{ViT} for its adaptation to images. The ViT architecture takes a grid of non-overlapping contiguous image patches of resolution $N\times N$ as input. Typically, we use $N=16$ (``/16'') in this paper. The patches are then mapped to a sequence of patch embeddings by a trainable linear projection. Some extra learnable tokens are appended to the sequence, e.g., class token \texttt{[CLS]} \cite{ViT,bert} and part token \texttt{[PART]} \cite{AAformer}. The role of these tokens is to aggregate information from the patch sequence. \texttt{[CLS]} is proposed to learn a global representation for the input image and \texttt{[PART]} aims to extract part-level feature. There is no structural difference between \texttt{[CLS]} and \texttt{[PART]}, while PASS will make them 
learn different-level features during training.

\subsubsection{Student network \& Teacher network.} 
\begin{figure}[t]
\centering
\includegraphics[width=1.\linewidth]{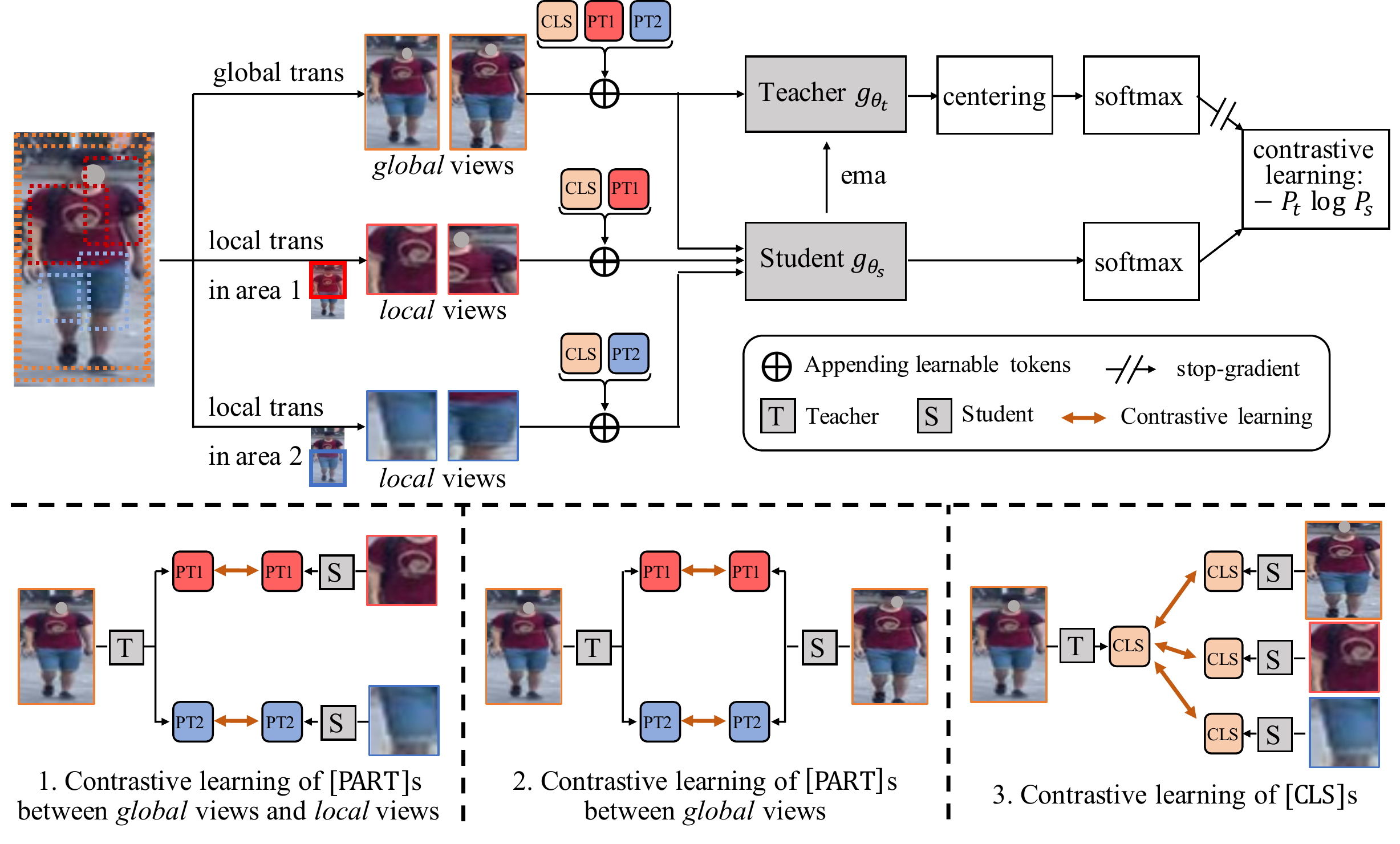}
\caption{The overview of PASS. We illustrate PASS in the case of $L=2$ for simplicity. The global random transformation (global trans) crops out the $global$ views from the whole images and the two local random transformations (local trans) crop out $local$ views from two local areas, respectively. The $local$ views from each local area are assigned a specific \texttt{[PART]}. The \texttt{[PART]}s of all local areas are also appended to the $global$ views.  \texttt{[CLS]} is also appended to all the views. In pre-training, all views pass through the student while only the $global$ views pass through the teacher. Each network predicts several $K$ dimensional features on \texttt{[CLS]} and \texttt{[PART]}s by the projection heads, and the features are normalized with a temperature softmax over the feature dimension. Then the similarities between the same \texttt{[PART]}/\texttt{[CLS]} output by student and teacher are measured with cross-entropy losses. We apply a stop-gradient operator on the teacher to propagate gradients only through the student. The teacher parameters are updated with an exponential moving average (ema) of the student parameters}
\label{fig:overview}
\end{figure}

The framework used for this work, PASS, shares a similar overall structure as the popular self-supervised approach, DINO \cite{DINO}. The overview of PASS is illustrated in Figure~\ref{fig:overview}, where the learning paradigm of knowledge distillation is used. PASS contains a student network and a teacher network, and they share the same architecture. PASS trains the student $f_{\theta_s}$ to match the output of the teacher $f_{\theta_t}$, which are parameterized by $\theta_s$ and $\theta_t$ respectively. Given an input image $x$, both networks predict probability distributions of $K$ dimensions on the appended learnable tokens (i.e., \texttt{[PART]} and \texttt{[CLS]}) by projection heads. The probability $P$ is obtained by normalizing the output of the network $f$ with a softmax function. Specifically, 
\begin{equation}
    P_{s}^g(x)^{(t)}=\frac{\exp \left(f_{\theta_{s}}^{cls}(x)^{(t)} / \tau_{s}\right)}{\sum_{k=1}^{K} \exp \left(f_{\theta_{s}}^{cls}(x)^{(k)} / \tau_{s}\right)},
\end{equation}
where $f_{\theta_{s}}^{cls}(x)$ is the predicted distribution on \texttt{[CLS]}, and $t,k$ are the indexes of vector components. $\tau_s>0$ is a temperature parameter that controls the sharpness of the output distribution. The  similar formulas hold for $P_{t}^g$, $P_{s}^{l_i}$, $P_{t}^{l_i}$, where $l_i$ means predicting on the $i$th \texttt{[PART]}.

The teacher $f_{\theta_t}$ is not pre-defined but built by means of the past iterations of the student $f_{\theta_s}$. We freeze the teacher network over an epoch and use an exponential moving average on the student weights \cite{DINO,MoCo_v1}. The update rule is:
\begin{equation}
\theta_t \gets \lambda\theta_t+(1-\lambda)\theta_s,
\end{equation}
with $\lambda$ following a cosine schedule from $0.996$ to $1$ during training. The output of the teacher network is centered with a mean calculated over the batch to avoid collapse \cite{DINO}.

\subsection{Part-Aware Self-Supervised Pre-training}

Given an input image $x$, PASS constructs a set of different views which includes different distorted views, or crops, of $x$. This set contains $M$ $global$ views, $x_m^g$, randomly cropped from the whole image, and $L\times J$ $local$ views of smaller resolution randomly cropped from $L$ local areas of the image,  where the $j$th cropped view of the $i$th local area is denoted as $x_j^{l_i}, (i\in\{1,...,L\}, j\in\{1,...,J\})$.

All views are passed through the student while only the $global$ views are passed through the teacher. For a $global$ view, all learnable tokens, i.e., \texttt{[CLS]} and \texttt{[PART]}s, are appended to the sequence of patch embeddings. While for a $local$ view cropped from the $i$th local area, only \texttt{[CLS]} and the $i$th part token \texttt{[PART]}$_i$ are appended to the patch embeddings. The projection heads are added to these extra learnable tokens in both teacher and student, and output the predicted probability distributions $P$. Specifically, given an input image $x$, its $m$th $global$ view $x_m^g$ is appended with \texttt{[CLS]} and all the \texttt{[PART]}s, and is passed through both the teacher and student. The predicted distributions by teacher and student are $P_t^g(x_m^g),~\{P_t^{l_i}(x_m^g)|i\in\{1,...,L\}\}$ and $P_s^g(x_m^g),~\{P_s^{l_i}(x_m^g)|i\in\{1,...,L\}\}$, respectively. Its $local$ view $x_j^{l_i}$ is only appended with \texttt{[CLS]} and \texttt{[PART]}$_i$, and then is only passed through the student. The student predicts two probability distributions on \texttt{[CLS]} and \texttt{[PART]}$_i$: $P_s^g(x_j^{l_i})$ and $P_s^{l_i}(x_j^{l_i})$.


PASS learns to match the predictions of the same learnable tokens by cross-entropy loss. Specifically, \textbf{for \texttt{[PART]}}$_i$, the student outputs two types of predictions:  (1) the prediction on \texttt{[PART]}$_i$ appended to $x_j^{l_i}$: $P_s^{l_i}(x_j^{l_i})$. (2) the prediction on \texttt{[PART]}$_i$ appended to $x_m^g$: $P_s^{l_i}(x_m^g)$. While the teacher only predicts the latter one and outputs $P_t^{l_i}(x_m^g)$. The student learns to match the output distributions of teacher by minimizing the cross-entropy loss w.r.t. the parameters of the student $\theta_s$:
\begin{equation}\label{e3}
    \min_{\theta_s}\left\{\sum_{m=1}^M\sum_{j=1}^JH\left(P_t^{l_i}\left(x_{m}^g\right),P_s^{l_i}\left(x_j^{l_i}\right)\right)+\sum_{m_1{=1}}^M\sum_{m_2{=1}}^M H\left(P_t^{l_i}\left(x_{m_1}^g\right), P_s^{l_i}\left(x_{m_2}^g\right)\right) \right\}
\end{equation}
where $m_1\neq m_2$ and $H(a,b)=-a\log b$. The two terms are corresponding to the contrastive learning 1 and 2 in the second row of Figure~\ref{fig:overview}. PASS applies this optimization to all the \texttt{[PART]}s. That is, any $i$ in $\{1,...,L\}$ is applicable in Eq.~\ref{e3}. 

In the student, the $local$ views assigned to each \texttt{[PART]} are randomly cropped from a specific local area, thus the \texttt{[PART]}s can focus on different areas. PASS uses the student to update the teacher, which can guarantee \texttt{[PART]}s in teacher also focus on different local areas and learns fine-grained information. In pre-training, the student learns to match the output of the teacher on the same \texttt{[PART]},
which can guarantee each \texttt{[PART]} learns a robust local representation from the $local$ views cropped in its corresponding local area.   

\textbf{For the \texttt{[CLS]}}, the predicted distributions on all the views are matched by:

\begin{equation}
\begin{aligned}
    \min_{\theta_s}\left\{
    \sum_{m{=1}}^M\sum_{i=1}^L\sum_{j=1}^JH\left(P_t^{g}\left(x_{m}^g\right), P_s^{g}\left(x_j^{l_i}\right)\right)+\sum_{m_1\atop{=1}}^M\sum_{m_2\atop{=1}}^M H\left(P_t^{g}\left(x_{m_1}^g\right), P_s^{g}\left(x_{m_2}^g\right)\right)
     \right\}
\end{aligned}
\end{equation}
where $m_1\neq m_2$. This formula makes the \texttt{[CLS]} learn the ``local-to-global'' correspondences and robust to the difficult scenarios, e.g., occlusion and incorrect detection, which are also why we still append \texttt{[CLS]} to the $local$ views.


\subsection{Fine-tuning}
 Pre-trained by PASS, the \texttt{[CLS]} of ViT backbone has the capability to  learn the global description of the input image, and the \texttt{[PART]}s can automatically focus on local areas and extract local representations. Therefore, in fine-tuning, we only need to append the \texttt{[CLS]} and all the \texttt{[PART]}s to the embedding sequence of the input image, and they can automatically focus on different-level features (we do not fixed them in fine-tuning).  
 The teacher network is used for fine-tuning.

\subsubsection{Supervised ReID.} We concatenate the output \texttt{[CLS]} and the mean of output \texttt{[PART]}s, denoted by \texttt{[$\overline{\texttt{Part}}$]}, as the representation of the input image. The ReID head \cite{BOT} is attached to the concatenated feature. Most of the training hyper-parameters are borrowed from the baseline of TransReID \cite{transreid}. That is, \textbf{none} of the overlapping patch embedding, jigsaw patch module, or side information embedding is used in our fine-tuning. The commonly used cross-entropy loss and triplet loss with hard sample mining \cite{triplet_loss} are adopted to train our model, of which the cross-entropy loss is calculated as:

\begin{equation}
    L_{cls}=-\log \mathcal{P}(\texttt{[CLS]}\textcircled{c}\texttt{[}\overline{\texttt{Part}}\texttt{]}),
\end{equation}
where $\mathcal{P}({\mathbf{x}})$ is the probability of $\mathbf{x}$ belonging to its ground truth identity and \textcircled{c} means concatenating.

The triplet loss is calculated as:
\begin{equation}
    L_{tri}=[d_p-d_n+\alpha]_{+},
\end{equation}
where $d_p$ and $d_n$ are feature distances from positive pair and negative pair, respectively. $\alpha$ is the margin of triplet loss. 
$[\cdot]_+$ equals to $max(\cdot, 0)$. Therefore, the overall objective function for our model is:
\begin{equation}
    L=L_{cls}+L_{tri}.
\end{equation}

In the testing phase, \texttt{[CLS]} and \texttt{[$\overline{\texttt{Part}}$]} are concatenated to represent a person image. 

\subsubsection{UDA/USL ReID.} We follow most of the settings in C-Contrast \cite{c_contrast} to conduct our experiments on UDA/USL ReID.  Before training in an unsupervised manner on target datasets, UDA ReID need to be first pre-trained on source datasets while USL ReID does not need. Apart from this, UDA ReID and USL ReID share the same training settings. First, all the training images pass through the network to obtain the training data features. Then, clustering is conducted on these features to generate pseudo labels. By averaging the features with the same pseudo labels, the cluster prototypes are obtained. Next in training, the contrastive loss between the output features and the cluster prototypes are computed to optimize the network. More details can be found in \cite{c_contrast}.

\section{Experiments}

\subsection{Implementation Details}

\hspace{1.5em}\textbf{Datasets.} LUPerson \cite{LUPerson}, which contains 4.18M unlabeled human images collected from $50,534$ online videos, are used for pre-training. To evaluate the performance on ReID tasks, we conduct the fine-tuning experiments on two widely used benchmarks, i.e., Market-1501 \cite{Market1501} and MSMT17 \cite{MSMT17}, which contain $32,668$ images of $1501$ identities and $126,441$ images of $4,101$ identities, respectively. In fine-tuning, the images are resized to $256\times 128$ unless mentioned otherwise. Following common practices, the cumulative matching characteristics (CMC) and the mean average precision (mAP) are used for evaluation.

\textbf{Pre-training.} For pre-training on LUPerson with PASS, the model is trained on  $8\times$A100 GPUs for 100 epochs, which costs about 60 hours for ViT-S and 120 hours for ViT-B. The divided local areas are with overlap to guarantee that the $local$ view can be cropped from anywhere in the image. The $global$ views are resized to $256\times 128$ and the $local$ views are resized to $96\times 48$. Similar to DINO, we set $M=2$ and $J = \lceil \frac{9}{L} \rceil$, e.g., $J=3$ when $L=3$ and $J=5$ when $L=2$.

\textbf{Supervised ReID.} To fine-tune the pre-trained Transformer, we use most of the training strategies of the baseline in TransReID \cite{transreid}, which means none of the overlapping patch embedding, jigsaw patch module, or side information embedding is used here. We set the learning rate to $lr=0.0004\times \frac{batchsize}{64}$ and warm up the model by 20 epochs \cite{reid_ssl}. The $\alpha$ in triplet loss is set to 0.3.

\textbf{UDA/USL ReID.} The ViTs are trained for 50 epochs and SGD is used. The initial learning rate is 3.5e-4 and is multiplied by 0.1 every 20 epochs. Each mini-batch contains 256 images of 32 persons, i.e., each ID contains 8 images. 

\subsection{Comparison with State-of-the-Art Methods}

\setlength{\tabcolsep}{4pt}
\begin{table}
\begin{center}
\caption{Comparison with the state-of-the-art methods of supervised ReID. Methods in the 1st group are pre-trained on ImageNet. Methods in the 2nd group are self-supervised methods pre-trained on LUPerson. Most of the self-supervised methods share the same fine-tuning settings as ours. The last group is our method. The results with underline are the best in their groups. ``TransReID$^-$'' means side information and overlapping patches are removed for a fair comparison}
\label{table:supervised_sota}
\begin{tabular}{llcccc}
\hline\noalign{\smallskip}
\multirow{2}*{Methods}&\multirow{2}*{Backbone}&\multicolumn{2}{c}{Market1501}&\multicolumn{2}{c}{MSMT17} \\ \cline{3-6}
~&~&mAP&Rank-1&mAP&Rank-1 \\ \noalign{\smallskip} \hline
 BOT \cite{BOT} & R50 & 85.9 & 94.5 & 50.2 & 74.1 \\ 
 MGN \cite{MGN} & R50$\uparrow$384 & 87.5 & 95.1 & 63.7 & \underline{85.1}\\
 SCSN \cite{SCSN} & R50$\uparrow$384 & \underline{88.5} & \underline{95.7} & 58.5 & 83.8 \\
 ABDNet \cite{ABD-Net} & R50$\uparrow$384 & 88.3 & 95.6 & 60.8 & 82.3 \\
 AAformer \cite{AAformer} & ViT-B$\uparrow$384& 87.7 & 95.4 & 63.2 & 83.6 \\
 TransReID$^-$ \cite{transreid} &ViT-B& 87.4 & 94.6 & 63.6 & 82.5 \\
 TransReID$^-$ \cite{transreid} &ViT-B$\uparrow$384 & 87.6 & 94.6 & \underline{65.8} & 84.4 \\
 \hline
 MoCoV2 \cite{MoCo_v2} & ViT-S & 72.1 & 87.6 & 27.8 & 47.4 \\
 MoCoV2 \cite{LUPerson} & MGN$\uparrow$384 & 91.0 & \underline{96.4} & 65.7 & 85.5 \\
 MoCoV3 \cite{MoCo_v3} & ViT-S & 82.2 & 92.1 & 47.4 & 70.3 \\
 MoBY \cite{MoBY} & ViT-S & 84.0 & 92.9 & 50.0 & 73.2 \\
 DINO \cite{DINO} & ViT-S & 90.3 & 95.4 & 64.2 & 83.4 \\
 DINO\,+\,CFS \cite{reid_ssl} & ViT-S & 91.0 & 96.0 & 66.1 & 84.6 \\
 DINO\,+\,CFS \cite{reid_ssl} & ViT-S$\uparrow$384 & \underline{91.5} & 96.0 & \underline{68.8} & \underline{86.1} \\
\hline
PASS (ours) & ViT-S & 92.2 & 96.3 & 69.1 & 86.5 \\
PASS (ours) & ViT-S$\uparrow$384 & 92.6 & 96.8 & 71.7 & 87.9 \\
PASS (ours) & ViT-B & 93.0 & 96.8 & 71.8 & 88.2 \\
PASS (ours) & ViT-B$\uparrow$384 & \textbf{93.3} & \textbf{96.9} & \textbf{74.3} & \textbf{89.7} \\
\hline
\end{tabular}
\end{center}
\end{table}
\setlength{\tabcolsep}{1.4pt}

\setlength{\tabcolsep}{4pt}
\begin{table}
\begin{center}
\caption{Comparison with the state-of-the-art methods of UDA ReID. Methods in the 1st group are pre-trained on ImageNet. Methods in the 2nd group are self-supervised methods pre-trained on LUPerson, and are fine-tuned with C-Contrast \cite{c_contrast}}
\label{table:uda_sota}
\begin{tabular}{llcccc}
\hline\noalign{\smallskip}
\multirow{2}*{Methods}&\multirow{2}*{Backbone}&\multicolumn{2}{c}{MSMT2Market}&\multicolumn{2}{c}{Market2MSMT} \\ \cline{3-6}
~&~&mAP&Rank-1&mAP&Rank-1 \\ \noalign{\smallskip} \hline
DG-Net++ \cite{DGNet} & R50 & 64.6 & 83.1 & 22.1 & 48.4 \\
MMT \cite{MMT} & R50 & 75.6 & 83.9 & 24.0 & 50.1 \\
SPCL \cite{SPCL} & R50 & 77.5 & 89.7 & 26.8 & 53.7 \\
MCRN \cite{MCRN} & R50 & - &- & 32.8 & \underline{64.4} \\
C-Contrast \cite{c_contrast} & R50 & \underline{82.4} & \underline{92.5} & \underline{33.4} & {60.5} \\
\hline
MoCoV2 \cite{MoCo_v2} & R50 & 85.1 & 94.4 & 28.3 & 53.8 \\
DINO \cite{DINO} & ViT-S & 88.5 & 95.0 & 43.9 & 67.7 \\
DINO\,+\,CFS \cite{reid_ssl} & ViT-S & \underline{89.4} & \underline{95.4} & \underline{47.4} & \underline{70.8} \\
\hline
PASS (ours) & ViT-S & \textbf{90.2} & \textbf{95.8} & \textbf{49.1} & \textbf{72.7} \\
\hline
\end{tabular}
\end{center}
\end{table}
\setlength{\tabcolsep}{1.4pt}

\subsubsection{Supervised ReID.} We compare PASS to some outstanding state-of-the-art methods on supervised ReID in Table~\ref{table:supervised_sota}. Compared with the methods pre-trained on ImageNet, our method significantly outperforms the existing best methods without adding any complex module to the backbone network, e.g., our ViT-B$\uparrow$384 obtains 93.3\%/74.3\% mAP and 96.9\%/89.7\% Rank-1 accuracy on Market1501/MSMT17, outperforming the state-of-the-art results by 4.8\%/8.5\% and 1.2\%/4.6\%, respectively. It is noted that AAformer \cite{AAformer} and TransReID are pre-trained on $ImageNet$-$21K$ which is much larger than LUPerson. 

Compared with the self-supervised methods pre-trained on LUPerson, PASS also shows considerable superiority, e.g., our ViT-S$\uparrow$384 obtains 92.6\%/71.7\% mAP and 96.8\%/87.9\% Rank-1 accuracy on Market1501/MSMT17 datasets,  surpassing the existing best method (CFS \cite{reid_ssl}) by 1.1\%/2.9\% and 0.8\%/1.8\%, respectively. DINO can be regarded as the baseline of our method, and the results validate the remarkable effectiveness of using \texttt{[PART]} to offer fine-grained local information in PASS. Besides, Table~\ref{table:supervised_sota} also shows that self-supervised pre-training on LUPerson is much more effective than supervised pre-training on ImageNet. The ICS module in \cite{reid_ssl} adds extra IBN \cite{IBN} layers to the ViT backbone and increases the overall complexity, thus is not compared here for fairness.

\subsubsection{UDA ReID.} Some latest UDA-ReID methods are compared in the Table~\ref{table:uda_sota}. Our ViT-S outperforms the existing best method by considerable margins. Specifically, ViT-S obtains 90.2\%/49.1\% (+0.8\%/+1.7\%) mAP  and 95.8\%/72.7\% (+ 0.4\%/+1.9\%) Rank-1 accuracy on MSMT17 $\rightarrow$ Market1501 and  Market1501 $\rightarrow$ MSMT17 respectively, which are already comparable to many supervised methods. Besides, PASS also surpasses its baseline method (DINO) by large margins, which validates local details are also vital for unsupervised learning in ReID.

\subsubsection{USL ReID.} We list some of the latest USL-ReID methods in Table~\ref{table:usl_sota}, which shows our method outperforms the existing best method by 0.7\% (94.9\% \emph{vs.} 94.2\%) and 0.6\% (67.0\% \emph{vs.} 66.4\%) on Rank-1 accuracy on Market and MSMT17, respectively. The results validate PASS can provide a better initialization to the ViT model on USL ReID.

\setlength{\tabcolsep}{4pt}
\begin{table}
\begin{center}
\caption{Comparison with the state-of-the-art methods of USL ReID. Methods in the 1st group are pre-trained on ImageNet. Methods in the 2nd group are self-supervised methods pre-trained on LUPerson, and are fine-tuned with C-Contrast \cite{c_contrast}}
\label{table:usl_sota}
\begin{tabular}{llcccc}
\hline\noalign{\smallskip}
\multirow{3}*{Methods}&\multirow{2}*{Backbone}&\multicolumn{2}{c}{Market1501}&\multicolumn{2}{c}{MSMT17} \\ \cline{3-6}
~&~&mAP&Rank-1&mAP&Rank-1 \\ \noalign{\smallskip} \hline
MMCL \cite{MMCL} & R50 & 45.5 & 80.3 & 11.2 & 35.4 \\
HCT \cite{HCT} & R50 & 56.4 & 80.0 & - & - \\
IICS \cite{IICS} & R50 & 72.9 & 89.5 & 26.9 & 52.4 \\
MCRN \cite{MCRN} & R50 & 80.8 & 92.5 & 31.2 & \underline{63.6} \\
C-Contrast \cite{c_contrast} & R50 & \underline{82.6} & \underline{93.0} & \underline{33.1} & 63.3  \\
\hline
MoCoV2 \cite{MoCo_v2} & R50 & 84.0 & 93.4 & 31.4 & 58.8 \\
DINO \cite{DINO} & ViT-S &87.8 & \underline{94.4} & 38.4 & 63.8 \\
DINO\,+\,CFS \cite{reid_ssl} & ViT-S & \underline{88.2} & 94.2 & \underline{40.9} & \underline{66.4} \\
\hline
PASS (ours) & ViT-S & \textbf{88.5} & \textbf{94.9} & \textbf{41.0} & \textbf{67.0} \\
\hline
\end{tabular}
\end{center}
\end{table}
\setlength{\tabcolsep}{1.4pt}

\subsection{Ablation Studies}

\subsubsection{The choices of $L$.} We first investigate the most suitable division strategy for PASS to divide the person images into several local areas. A $local$ view accounts for up to 40\% of the whole image, thus when $L=2$, each local area should occupy 70\% of the image (the same width as the original image and 70\% height) to guarantee that the $local$ view can be cropped from almost anywhere in the image. For the same reason, when $L=3$, each local area occupies 50\% of the image. We also follow MGN \cite{MGN} to conduct the ablation experiments which use $L=\{2,3\}$, where two kinds of division granularity are used at the same time. The results in Table~\ref{table:ablation_k} show that increasing the number of local areas does not always bring positive feedback, and PASS with $L=3$ obtains the best performance. Therefore, we recommend dividing the image into 3 local areas with each area occupying 50\% of the image. 


\setlength{\tabcolsep}{4pt}
\begin{table}
\begin{center}
\caption{Ablation studies about the number of divided local areas $L$ on supervised ReID}
\label{table:ablation_k}
\begin{tabular}{cccccc}
\hline\noalign{\smallskip}
\multirow{2}*{L}&\multirow{2}*{Backbone}&\multicolumn{2}{c}{Market1501}&\multicolumn{2}{c}{MSMT17} \\ \cline{3-6}
~&~&mAP&Rank-1&mAP&Rank-1 \\ \noalign{\smallskip} \hline
 2 & ViT-S & 91.7 & 96.1 & 67.7 & 85.6 \\
 3 & ViT-S & \textbf{92.2} & \textbf{96.3} & \textbf{69.1} & 86.5 \\
 4 & ViT-S & {91.9} & {96.1} & {68.1} & {86.0} \\
 5 & ViT-S & {90.9} & {95.8} & {67.3} & {85.4} \\
 2,3 & ViT-S & 92.0 & 96.3 & 68.7 & \textbf{86.6} \\
\hline
\end{tabular}
\end{center}
\end{table}
\vspace{-30pt}
\setlength{\tabcolsep}{1.4pt}

\subsubsection{Feature fusion.} Next, we conduct experiments to investigate how to effectively integrate the output global feature and local features on supervised/UDA/USL ReID, which are shown in Table~\ref{table:ablation_feature} and Table~\ref{table:ablation_feature_UDA}, respectively. One way is directly concatenating all these features, which is used by TransReID \cite{transreid}. However, this way increases the dimension of feature embedding several times and cannot obtain the best performance in our experiments. We also conduct the ablation study which uses the mean feature or concatenates \texttt{[CLS]} and \texttt{[$\overline{\texttt{Part}}$]}. The results show concatenating \texttt{[CLS]} and \texttt{[$\overline{\texttt{Part}}$]} can obtain the best performance on supervised ReID. It is also worth noting that the strategy of ``mean feature'', which does not increase the feature dimension, already outperforms the existing state-of-the-art methods, some of which increase the feature dimension several times \cite{MGN,transreid}, by considerable margins. On UDA/USL ReID, MSMT17 dataset prefers the ``mean feature'' while Market1501 prefers concatenating \texttt{[CLS]} and \texttt{[$\overline{\texttt{Part}}$]}, which may be related to their data distributions. 

\setlength{\tabcolsep}{4pt}
\begin{table}[t]
\begin{center}
\caption{Ablation studies about feature fusion on supervised ReID. \textcircled{c} means concatenating. $C$ is the number of channels. The first row concatenates all these features. The second row uses the mean feature. The third row concatenates \texttt{[CLS]} and \texttt{[$\overline{\texttt{Part}}$]}}
\label{table:ablation_feature}
\begin{tabular}{ccccccc}
\hline\noalign{\smallskip}
\multirow{2}*{Method}&\multirow{2}*{Dim}&\multirow{2}*{Backbone}&\multicolumn{2}{c}{Market1501}&\multicolumn{2}{c}{MSMT17} \\ \cline{4-7}
~&~&~&mAP&Rank-1&mAP&Rank-1 \\ \noalign{\smallskip} \hline
 \multirow{2}*{\texttt{[CLS]}\textcircled{c}$\frac{\texttt{[PART]}_1}{L}$\textcircled{c}...$\frac{\texttt{[PART]}_L}{L}$} & \multirow{2}*{(L+1)$\times$C} &ViT-S & 90.8 & 96.1 & 67.7 & 85.2 \\
 ~&~& ViT-B & 92.2 & 96.3 & 71.3 & 87.4  \\ \hline
 \multirow{2}*{$\frac{1}{2}$(\texttt{[CLS]}+\texttt{[$\overline{\texttt{Part}}$]})} & \multirow{2}*{C} &ViT-S & 92.0 & \textbf{96.3} & 68.7 & 86.3 \\
 ~&~&ViT-B& 92.5 & 96.7 & 71.5 & 87.8 \\ \hline
 \multirow{2}*{\texttt{[CLS]}\textcircled{c}\texttt{[$\overline{\texttt{Part}}$]}} & \multirow{2}*{2C} &ViT-S & \textbf{92.2} & \textbf{96.3} & \textbf{69.1} & \textbf{86.5} \\
 ~&~&ViT-B& \textbf{93.0} & \textbf{96.8} & \textbf{71.8} & \textbf{88.2} \\
\hline
\end{tabular}
\end{center}
\end{table}
\setlength{\tabcolsep}{1.4pt}

\setlength{\tabcolsep}{4pt}
\begin{table}[t]
\begin{center}
\caption{Ablation studies about feature fusion on UDA/USL ReID. The backbone is ViT-S and \textcircled{c} means concatenating. The first row concatenates all these features. The second row uses the mean feature. The third row concatenates \texttt{[CLS]} and \texttt{[$\overline{\texttt{Part}}$]}}
\label{table:ablation_feature_UDA}
\begin{tabular}{ccccccccc}
\hline\noalign{\smallskip}
\multirow{3}*{Method}& \multicolumn{4}{c}{UDA reID} & \multicolumn{4}{c}{USL reID} \\ \cline{2-9}
~&\multicolumn{2}{c}{MS2MA}&\multicolumn{2}{c}{MA2MS}&\multicolumn{2}{c}{Market1501}&\multicolumn{2}{c}{MSMT17} \\ \cline{2-9}
~&mAP&R-1&mAP&R-1&mAP&R-1&mAP&R-1 \\ \noalign{\smallskip} \hline \noalign{\smallskip}
 \texttt{[CLS]}\textcircled{c}$\frac{\texttt{[PART]}_1}{L}$\textcircled{c}...$\frac{\texttt{[PART]}_L}{L}$ & 90.0 & 95.6 & 47.1 & 71.1 &  88.4 & 94.6 & 39.4 & 64.9 \\ \noalign{\smallskip} \hline \noalign{\smallskip}
 $\frac{1}{2}$(\texttt{[CLS]}+\texttt{[$\overline{\texttt{Part}}$]}) & 89.8 & 95.5 & \textbf{49.1} & \textbf{72.7} & \textbf{88.6} & 94.6 & \textbf{41.0} & \textbf{67.0} \\ \noalign{\smallskip} \hline \noalign{\smallskip} 
 \texttt{[CLS]}\textcircled{c}\texttt{[$\overline{\texttt{Part}}$]} & \textbf{90.2} & \textbf{95.8} & 44.9 & 69.4 & 88.5 & \textbf{94.9} & 36.6 & 62.5   \\ \noalign{\smallskip} \hline
\end{tabular}
\end{center}
\end{table}
\setlength{\tabcolsep}{1.4pt}

\begin{figure}[h!]
\centering
\includegraphics[width=.9\linewidth,height=5cm]{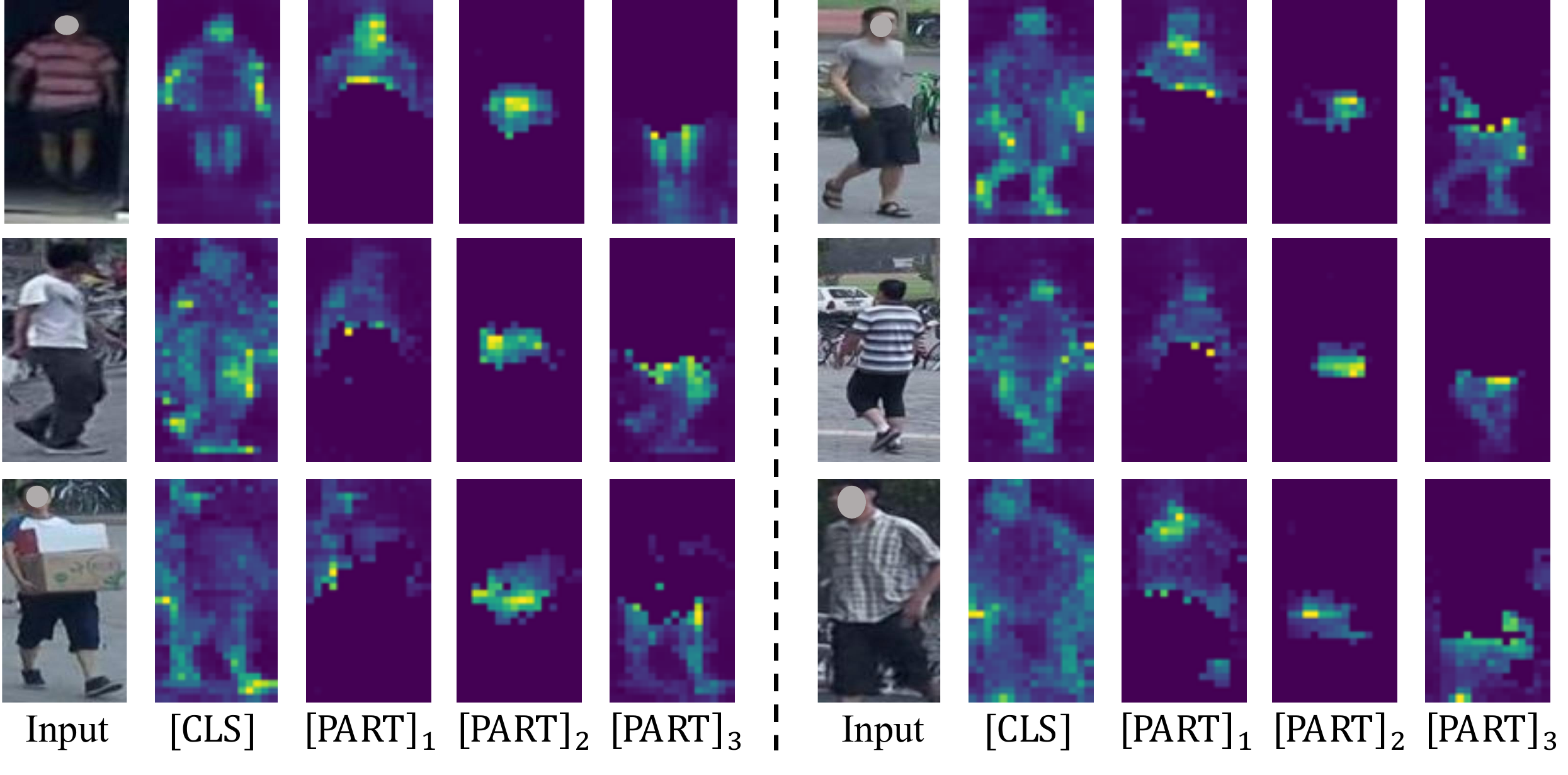}
\caption{Visualization of attention maps of \texttt{[CLS]} and \texttt{[PART]}s in the last self-attention layer of ViT pre-trained by PASS. For each \texttt{[PART]}, the patches that have the maximal similarity with it among all the \texttt{[PART]}s are visualized. The examples in the last row show the visualization on occlusion and bad detection scenarios}
\label{fig:visualization}
\end{figure}

\subsubsection{Visualization.} To further give an intuitive illustration of the effectiveness of PASS, we conduct visualization experiments to show the focus areas of \texttt{[CLS]} and \texttt{[PART]}s in the ViT backbone pre-trained by PASS. As illustrated in Figure~\ref{fig:visualization}, given an input image, \texttt{[CLS]} focuses on the whole images to extract the global feature, and the \texttt{[PART]}s focus on different local areas, i.e., upper-body, waist, and legs, to extract part-level features, which can obtain more fine-grained information. More satisfactory, \texttt{[PART]}s can locate on specific semantic parts rather than simply locating different positions. Take the bottom-left sample in Figure~\ref{fig:visualization} as an example, when a man carries a box and the box blocks his upper body, the first \texttt{[PART]} does not focus on the box but focus on the visible part of his upper body. This makes the learned feature more robust as the man may carry boxes with different colors in other images. This visualization also validates PASS can well handle the occlusion scenarios. It is worth noting that we do not add any extra module to the ViT backbone to achieve this, but only pre-train the ViT backbone with PASS.

\subsubsection{Ranking list.} 
\begin{figure}[h]
\centering
\includegraphics[width=.9\linewidth]{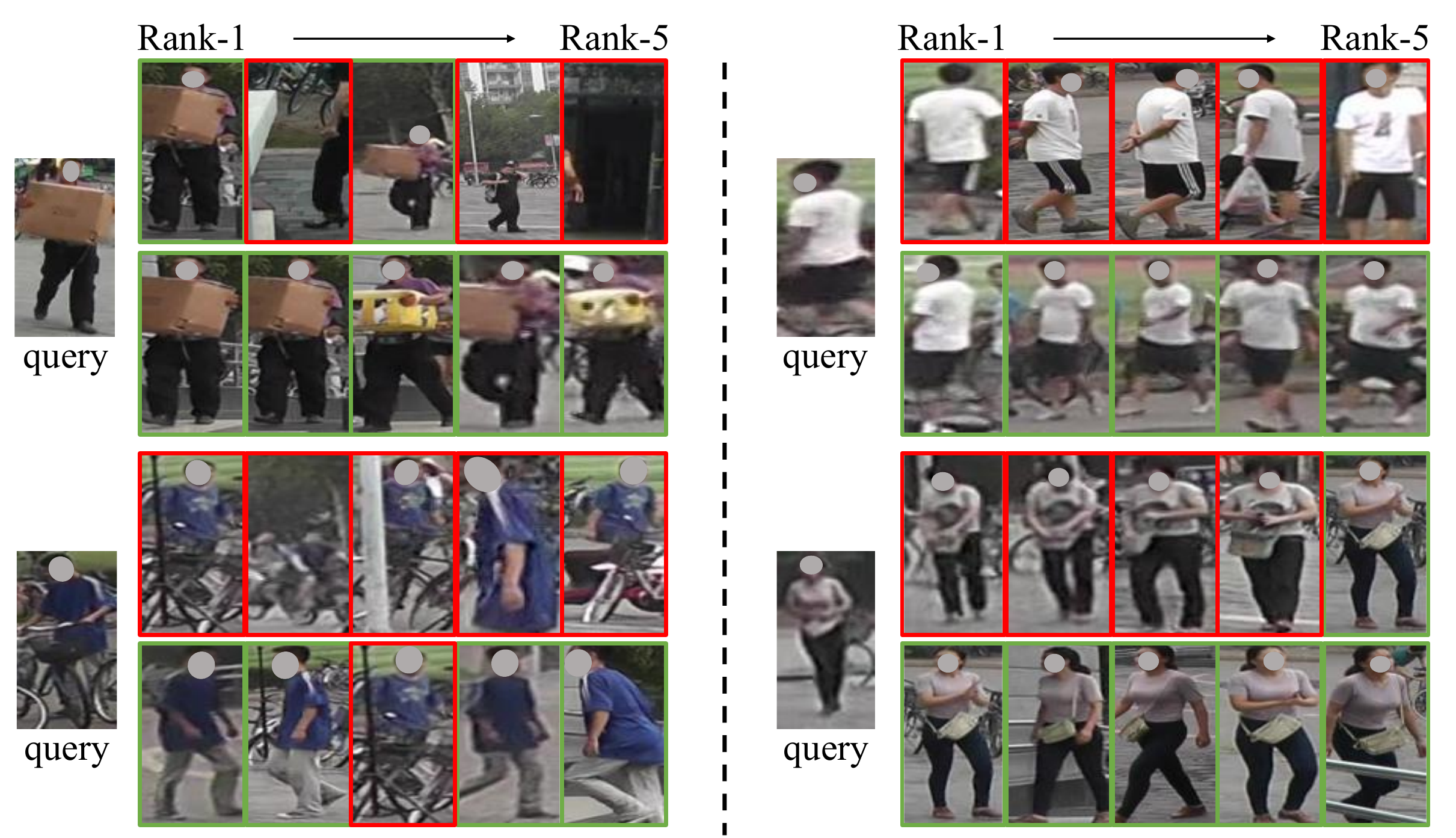}
\caption{The ranking lists of models pre-trained by DINO and PASS. For each query, the ranking list of the first row is from DINO and that of the second row is from PASS. The results show the PASS is largely superior to DINO in finding the fine-grained information of local details}
\label{fig:ranking_list}
\end{figure}

Finally, we compare the ranking results of ViT pre-trained by PASS with DINO (baseline) in Figure~\ref{fig:ranking_list}. \emph{The first example shows even the person takes different boxes with different colors, our model can still find him in the massive images, which validates our model is not affected by the obstructions and focuses on the details of visible parts.} The other samples show the strong ability of our model in discriminating the extremely similar samples through the identifiable tiny clues, e.g., patterns on clothes. All these samples validate the superiority of PASS over the baseline method DINO.

\section{Conclusion}
In this paper, we propose a ReID-specific self-supervised pre-training method, Part-Aware Self-Supervised pre-training (PASS). Pre-trained by PASS, the models can automatically extract part-level features from the input images and offer  more fine-grained information. Experimental results validate the powerful performance of PASS in both supervised ReID and UDA/USL ReID.

\subsubsection{Acknowledgement}. This work was supported by National Key R\&D Program of China under Grant No.2021ZD0110403, National Natural Science Foundation of China (No.62002356, 61976210, 62002357, 62076235), Open Research Projects of Zhejiang Lab (No.2021KH0AB07). 


%
%
\bibliographystyle{splncs04}
\bibliography{egbib}
\end{document}